# Speech-Based Blood Pressure Estimation with Enhanced Optimization and Incremental Clustering

Vaishali Rajput[1,2][0000-0002-9399-0554], Preeti Mulay[1][0000-0002-4779-6726], Rajeev Raje[3]
1 Symbiosis Institute of Technology, Symbiosis International (Deemed University), Pune, India.
2 Vishwakarma Institute of Technology, Pune, India.
3 Indiana- Purdue University, Indiana, United States

*Contact Details: Vaishali Rajput (vaishali.kalyankar.phd2019@sitpune.edu.in)

**Abstract**

Blood Pressure (BP) estimation plays a pivotal role in diagnosing various health conditions, highlighting the need for innovative approaches to overcome conventional measurement challenges. Leveraging machine learning and speech signals, this study investigates accurate BP estimation with a focus on preprocessing, feature extraction, and real-time applications. An advanced clustering-based strategy, incorporating the k-means algorithm and the proposed Fact-Finding Instructor optimization algorithm, is introduced to enhance accuracy. The combined outcome of these clustering techniques enables robust BP estimation. Moreover, extending beyond these insights, this study delves into the dynamic realm of contemporary digital content consumption. Platforms like YouTube have emerged as influential spaces, presenting an array of videos that evoke diverse emotions. From heartwarming and amusing content to intense narratives, YouTube captures a spectrum of human experiences, influencing information access and emotional engagement. Within this context, this research investigates the interplay between YouTube videos and physiological responses, particularly Blood Pressure (BP) levels. By integrating advanced BP estimation techniques with the emotional dimensions of YouTube videos, this study enriches our understanding of how modern media environments intersect with health implications. Performance evaluation through metrics including Davies Bouldin score, Homogeneity, completeness, Jacquard similarity, Silhouette score, and Dunn's index demonstrates substantial enhancements, particularly with a 90% training percentage. This method offers promising potential for accurate BP estimation, contributing to the evolution of assessment methodologies and ultimately enhancing healthcare outcomes.

**Introduction:**

According to Kaur et al. (2019) a human disease is a particular aberrant state that has a detrimental effect on an organism's overall structure or function but is not instantly caused by an external injury. According to Gautam et al.(2019) there are four main groups of diseases that affect humans: infectious diseases, deficiency disorders, hereditary diseases, and physiological diseases. Diseases can also be categorized in other ways, such as communicable or non-communicable (Gautam et al.,

2020). Communicable diseases are those that can transfer from one person to another. According to Sharma et al. (2017), non-communicable diseases are those that cannot be passed from one person to another. Estimating Blood Pressure (BP) is essential for detecting several disorders, making it one of the most important health indicators. Invasive and non-invasive approaches are used to estimate Blood Pressure, with the invasive method providing a higher estimation accuracy but with its own difficulties and restrictions. According to the World Health Organization's (WHO) 2015 estimate, 9.4 million people worldwide pass away from high Blood Pressure (hypertension), and 25% of women and 30% of men have BP (Argha et al., 2021, Farki et al., 2021).

Following diabetes as the second most common cause of death from cardiovascular disease, hypertension is regarded as a silent killer disease because it has no symptoms. Most of the clinical settings routinely check the patient's Blood Pressure, and the same is done for elderly patients and those in the Intensive Care Unit (ICU). Regular Blood Pressure monitoring can help prevent diseases like heart failure, heart attacks, and stroke (Liu et al., 2017, Shahabi et al., 2015). Additionally, over time, hypertension harms human organs like the kidneys, eyes, and brains Farki et al., 2021). BP has an impact on the air pressure that builds up in the lungs. Additionally, Blood Pressure (BP) influences heart rate, and references to the relationship between heart rate and human speech recordings may be found in (Mesleh et al., 2012, Kim et al., 2004). As a result, it has been increasingly essential in recent years to investigate the BP using an auditory signal (Ankışhan, 2020). Machine learning is quite successful in classifying and predicting the acoustic signal. The temporal and frequency domains of the audio signal can be considered when classifying the audio (Song et al., 2012, Krizhevsky et al., 2017). The elimination of redundant information is the most important step for reducing computational complexity since preliminary processes like pre-processing and significant feature extraction increase prediction accuracy and cut down on calculation time. Removing unnecessary features, cutting processing time, and data augmentation are the three most important phases in the preparatory step for BP estimation. However, while considering real-time applications, the enhanced features raise the computation's complexity. As a result, the clustering-based approach's basic characteristics improve BP estimation accuracy (Farki et al., 2021)

**Synergistic Approach-Incremental Clustering with K-means and Fact Finding Algorithm:**
In this section, we present a synergistic approach that combines incremental clustering with the power of the k-means algorithm and the Fact Finding Instructor Optimization algorithm. This innovative combination allows for dynamic and real-time clustering of time series speech data for accurate Blood Pressure estimation. By leveraging the strengths of each algorithm, we achieve continuous updates and improved clustering accuracy, providing valuable insights for healthcare professionals in diagnosing, monitoring, and managing patients' Blood Pressure levels (Bagirov et al., 2011).

**Proposed Fact Finding Instructor based BP Estimation:**

The Fact-Finding Instructor Optimization algorithm is developed by combining the investigative skills of a fact finder in identifying suspects of criminal offenses with the expertise and knowledge of an instructor to enhance the performance of a trainee (Rao et al., 2016). By leveraging these characteristics, the algorithm aims to optimize the BP estimation process.

The fact-finding aspect of the algorithm enables it to identify relevant patterns and features within the speech signals that are indicative of BP levels. This investigative approach helps in accurately determining the suspect (i.e., the BP value) from the speech signal data. By incorporating the instructor's knowledge, the algorithm ensures that the trainee (i.e., the BP estimation model) performs optimally and achieves high accuracy.

By hybridizing the fact finding and instructor components, the proposed algorithm aims to overcome the limitations of existing BP estimation techniques. It is designed to provide a more efficient and accurate estimation of BP, enabling better disease diagnosis and effective management of hypertension.

In the proposed Fact-Finding Instructor-based BP estimation method, the collected speech signal undergoes preprocessing to remove any noise and artifacts. This is achieved by applying an adaptive filter, which helps enhance the quality of the signal. Once the preprocessing step is complete, various features are extracted from the speech signal. These features include amplitude, frequency, and width formant features, which provide information about the characteristics of the speech signal. Additionally, statistical features such as zero crossing, change in zero crossing, Haar Wavelet, pitch function, loudness function, entropy, Mel-LPC, variance, mean, and harmonic ratio are computed. These features collectively form a feature vector that represents the speech signal.

The next step in the proposed method involves clustering the BP using two algorithms: the K-means clustering algorithm and the proposed fact-finding instructor optimization algorithm. The K-means clustering algorithm groups similar data points together based on their similarity in feature space (Sinaga et al., 2020). It helps identify clusters or patterns within the extracted features that are indicative of different BP levels.

The fact-finding instructor optimization algorithm, which was introduced earlier, guides the clustering process by leveraging the characteristics of a fact finder and an instructor. This algorithm aims to find the optimal clustering solution by balancing intensification (focus on local best solutions) and diversification (exploration of the search space) phases. It considers the instructor's knowledge to enhance the accuracy and efficiency of the clustering process.

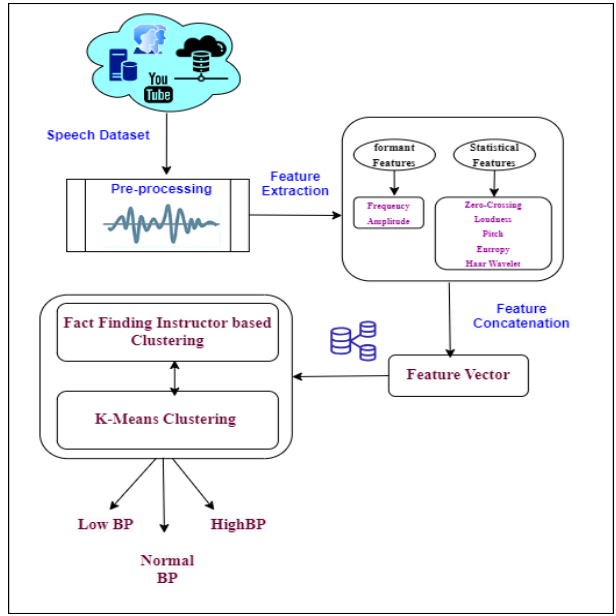

**Fig. 1:** Block Diagram of proposed Fact Finding Instructor based BP estimation.

Finally, the output generated by both clustering operations is combined by multiplying them together. This combined output serves as the estimation of the BP. By leveraging the clustering results and the optimization algorithm, the proposed method aims to provide an accurate estimation of BP based on the speech signal.

The proposed Fact-Finding Instructor-based BP estimation method is illustrated in Fig.1, which visually depicts the different stages of the algorithm and their interconnections.

Overall, this method utilizes preprocessing, feature extraction, clustering algorithms, and optimization techniques to estimate BP from speech signals. By integrating the fact finding and instructor components, it aims to enhance the accuracy and reliability of BP estimation.

**Feature Extraction:**

In the proposed clustering-based BP estimation method, the first step involves extracting features from the patient's input audio signal. These features are essential for capturing relevant information related to BP levels. Before feature extraction, a preprocessing stage is performed to remove any artifacts or noise present in the audio signal.

During feature extraction, both statistical-based features and formant features of the audio signal are considered. Statistical-based features include various measures that describe the statistical properties of the signal, such as mean, variance, entropy, and zero-crossing rate. These features provide information about the overall characteristics and distribution of the audio signal.

Formant features, on the other hand, capture specific properties related to the frequency content and resonance of the audio signal. Formants are distinct frequency bands that represent the resonant frequencies of the vocal tract during speech production. By extracting formant features, the algorithm can capture the unique patterns and characteristics in the speech signal that are relevant to BP estimation.

By combining statistical-based features and formant features, the algorithm aims to capture a comprehensive representation of the audio signal while reducing computational complexity. This selection of features helps to focus on the most informative aspects of the signal for BP estimation, without overwhelming the algorithm with unnecessary data.

The extraction of these features from the pre-processed audio signal forms the basis for subsequent stages of the clustering-based BP estimation process. By incorporating both statistical-based and formant features, the algorithm can effectively represent the speech signal and extract meaningful information for accurate BP estimation.

**Formant Features:**

In the proposed BP estimation method, formant features such as amplitude, width, and frequency are extracted from the patient's input audio signal. These formant features provide valuable information about the characteristics and properties of the speech signal, which can be correlated with BP levels.

**Statistical Features:**

In addition to the formant features, the proposed BP estimation method incorporates various statistical features like Zero Crossing, Entropy, Change in Zero Crossing, Haar Wavelet, Loudness, Pitch, Mel-LPC, Harmonic Ratio, Mean, *Variance* are extracted from the audio signal. These features are chosen to provide relevant information about the signal characteristics and contribute to the accurate estimation of BP.

**Motivation:**

The algorithm comprises a fact-finding phase that identifies audio signal features indicative of BP levels. Formant features, statistics, and pertinent data are extracted from speech signals. The subsequent chasing phase employs clustering to categorize these features, akin to a team pursuing leads. This process uncovers patterns and relationships, grouping data points into distinct BP categories. The collaboration between the fact-finding and chasing teams is mirrored in the interplay between feature extraction and clustering (Chou & Nguyen, 2020, Rao, 2016). The algorithm leverages instructor-like guidance to optimize feature extraction accuracy, enhancing its performance for managing hypertension. The instructor component ensures the optimal functioning of the BP estimation model. Integrating these elements bolsters the accuracy and reliability of BP estimations, contributing to more effective hypertension management. The instructor also enhances the exploration phase by sharing expertise with the estimation model. This collaboration amplifies the algorithm's capacity to explore the feature space comprehensively, identifying global optimal solutions for clustering BP classes in patients. This approach strikes a balance between exploration and exploitation, thereby sidestepping local optima and achieving more accurate estimations. The algorithm's advantages include rapid convergence due to an improved fact-finding phase, as well as comprehensive BP class clustering based on speech features. In sum, the proposed algorithm synergizes the instructor's expertise with the fact-finding team's insights to cluster BP classes. This hybrid approach enhances exploration, achieves a balanced trade-off, and ensures swift, accurate convergence. By harnessing speech features, the algorithm offers a holistic solution for BP estimation in patients.

**Mathematical Modeling of the Fact-Finding algorithm:**

The proposed BP estimation algorithm involves a two-team approach: Fact-finders and chasers. Fact-finders use their expertise to identify potential accurate BP estimation locations. They share their findings with the chasing team, who focus on these locations. The process is coordinated by headquarters, with Fact-finders guiding chasers and receiving updates. This collaboration ensures accurate identification through continuous feedback. The algorithm's core principle is illustrated in Fig.2.

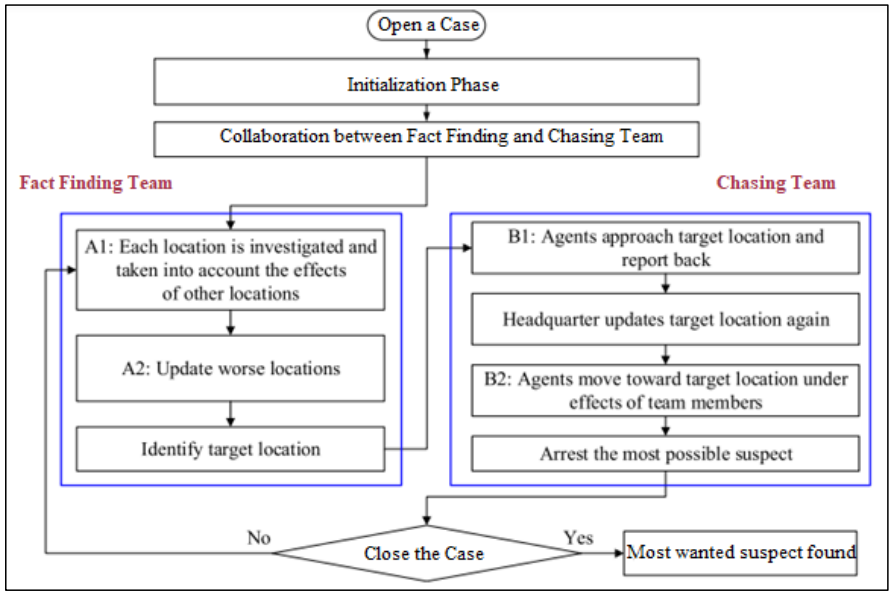

Fig.2: General principal of the proposed FFI algorithm

Let the fact-finding team is notated as $FF$ and the chasing team is notated as $CT$. The position of the suspect that needs to be investigated is notated as $D_{FF}^{\tau}$, which $\tau$ refers to the location of the suspect that ranges between $\tau = 1,2,..., X_{FF}$, here, the total suspect's location needs to be detected is referred as $X$. The location of the police corresponding to the chasing team is denoted as $D_{CT}^{\tau}$, which $\tau$ refers to $\tau = 1,2,..., X_{CT}$ and the total number of police in the chasing team is notated as $X$. In the proposed Fact Finding Instructor Optimization algorithm, the total number of suspect's location and police are the same and is notated as $X_{FF} = X_{CT} = X$ and is considered as the population size. The iteration of the proposed Fact Finding Instructor Optimization algorithm has been notated $i$ and its maximal value is indicated as $i_{max}$. The dimension of the parameter vector is notated as $Y$.

**i) Detection of Suspect's Location**: The possible location of the suspect's hiding place is analyzed in this stage and the first location of the suspect is denoted as $\left(D_{FF}^{\tau}\right)'$, and the general formulation is expressed as,

$$\left(D_{FF}^{\tau}\right)' = D_{FF}^{\tau} + A * \left[\frac{\sum_{1}^{B1} D_{FF}^{d}}{B1}\right] \quad (1)$$

where, $l = 1,2,.....Y$, the random number ranges between $[-1,1]$ is indicated as $A$, and $B1 \in \{1,2,.......a-1\}$. In this, the random parameter $A$ is expressed as $A = (rand - 0.5) * 2$, here the value of rand ranges between $[0,1]$. $D_{FF}^{d}$ is the location of the suspect location, the objective location (possible location) of the suspect is notated as $C_{FF}^{\tau}$, and the location of the new suspect is denoted as $\left(C_{FF}^{\tau}\right)'$. By evaluating the suspect's location based on the trial and error approach, the value of $B1$ is assumed as 2 and then the location is rewritten as,

$$\left(D_{FF}^{\tau}\right)' = D_{FF}^{\tau} + A * \left[D_{FF}^{d} - \frac{\left(D_{FF}^{ql} + D_{FF}^{pl}\right)}{2}\right] \quad (2)$$

where $q, p, and\ \tau$ are locations of the suspect $q\ and\ p$ are chosen randomly. $\{q, p, \tau\} \in \{1,2,......, X\}$

In the detection of the suspect location, the instructor's knowledge in instructing the trainee is incorporated for the detection of a more accurate hiding location of the suspect. The position of the instructor based on the trainee's score is expressed as,

**ii) Inquiry Direction**: In this stage, the probability of the suspect's hiding location is investigated, in which the best possible location is indicated as $e_{min}$ and the worst possible location is indicated as $e_{max}$. The best location is notated as $D_{min}$, and the probability of each suspect's location is:

$$prob(D_{FF}^{\tau}) = \frac{(e_{max} - C_{FF}^{\tau})}{e_{max} - e_{min}} \qquad (3)$$

Here, the location of the suspect is changed to increase the exploration area and hence the general formulation is expressed as in equation (3). The movement of the suspect location is affected by the best individual.

$$(D_{FF}^{\tau l})' = D_{min} + \sum_{1}^{B2} \beta * D_{FF}^{kl} \qquad (4)$$

where the individuals that affect the move are indicated as $B2$ and ranges between $B2 \in \{1,2,.....a-1\}; k = 1,2,....B2$; and $\beta = [-1,1]$ is considered as the coefficient of effectiveness. The value of $B2$ is equal to 3 and then the suspect's new location is evaluated as,

$$(D_{FF}^{\tau l})^2 = D_{min} + D_{FF}^{bl} + A5 * (D_{FF}^{ql} - D_{FF}^{pl}) \qquad (5)$$

where the random number with the range [0,1] is referred to as $A5$ and $(D_{FF}^{\tau l})^2$ is the second location of the suspect generated by the fact-finding team.

iii) **Action**: The fact-finding team provides the best location for the suspect to identify the suspect. The chasing team starts the suspect chasing in a coordinated fashion and the search agent named police $CT_\tau$ reaches the suspect's location hence the new location of the police agent is notated as,

$$(D_{CT}^{\tau l})'' = D_{CT}' + [A3 - 0.5] * 2 * \frac{\sum D_{CT}'}{A4} \qquad (6)$$

where the random number $A3$ ranges between [0,1] and the random number $A4$ ranges between [-1 to 1].

iv) **Fitness**: Fitness is evaluated for the attainment of the desired solution for solving optimization issues. In the proposed fact-finding optimization algorithm for clustering the BP of the patients, the accuracy of clustering is evaluated as a fitness function and is expressed as,

$$BP_{fit} = \frac{BP_{tp} + BP_{tn}}{BP_{tp} + BP_{tn} + BP_{fp} + BP_{fn}} \qquad (7)$$

where the true positive is notated as $BP_{tp}$ true negative is notated as $BP_{tn}$ false positive is notated as $BP_{fp}$ and false negative is notated as $BP_{fn}$ and the fitness function is notated as $BP_{fit}$.

Equation (6) increases the convergence rate of the algorithm by considering the investing characteristics of the search agents. It is mentioned that the instructor should possess' high knowledge to direct the chasing team to avoid trapping in the local optima. Hence, the final updated equation of the algorithm is given by,

$$D_{CT,Te}^{\tau l} = 0.5 \left[ D_{CT}^{\tau l} \right] + 0.5 \left[ D_{Te}^{\tau l} \right] \qquad (8)$$

Where $D_{Te}^{\tau l}$ is the location directed by the instructor through the knowledge gained by the teaching process and it is given as:

$$D_{Te}^{\tau l} = D_{Te}^{\tau l-1} + A1\left(D_{Te}^{\tau l-1} - T*\mu\right) \qquad (9)$$

Where the $D_{Te}^{\tau l-1}$ previous solution provided by the leaner, $T$ is the teaching factor and $\mu$ represents the mean.

Hence, by substituting equations (6) and (9) in equation (8) we get:

$$D_{CT,Te}^{\tau l} = 0.5\left[D_{CT}^{'} + [A3-0.5]*2*\frac{\sum D_{CT}^{'}}{A4}\right] + 0.5\left[D_{Te}^{\tau l-1} + A1\left(D_{Te}^{\tau l-1} - T*\mu\right)\right] \qquad (10)$$

**v) Termination**: The termination of the optimization process in the proposed Fact-Finding Instructor Optimization algorithm occurs under two conditions:

1. Proximity to the Objective Function: The algorithm stops when the solution acquired is close to the objective function. This means that the fitness or performance of the current solution is considered satisfactory, and further iterations are not necessary.
2. Maximum Iterations: The algorithm also terminates when the maximum specified number of iterations is reached. This is a predefined limit set at the beginning of the algorithm to control the computational resources and prevent excessive computation.

These termination conditions ensure that the optimization process in the Fact-Finding Instructor Optimization algorithm stops either when an acceptable solution is found (proximity to the objective function) or when the algorithm has reached the maximum allowed iterations.

By terminating the optimization process, the algorithm aims to strike a balance between computational efficiency and the quality of the obtained solution. It prevents unnecessary iterations while ensuring that the algorithm has sufficient opportunity to converge towards a satisfactory result. The mathematical model of the FFI is shown in Fig.3.

In the proposed Fact-Finding Instructor Optimization algorithm, the global best solution is determined using equation (10). This equation integrates the investigating characteristics of the instructor, along with the knowledge obtained through the learning characteristics of the instructor. The instructor's investigative nature broadens the search scope for the fact-finding and chasing teams, enhancing exploration in the solution space (Chou & Nguyen, 2020, Rao, 2016). This phase maximizes solution coverage, improves convergence, and refines BP class clustering. Furthermore, the instructor's wisdom prevents premature convergence, balancing exploration, and exploitation. This integration ensures high-quality solutions and boosts clustering performance. Ultimately, by capitalizing on investigative traits and instructor knowledge, the algorithm optimizes BP estimation accuracy.

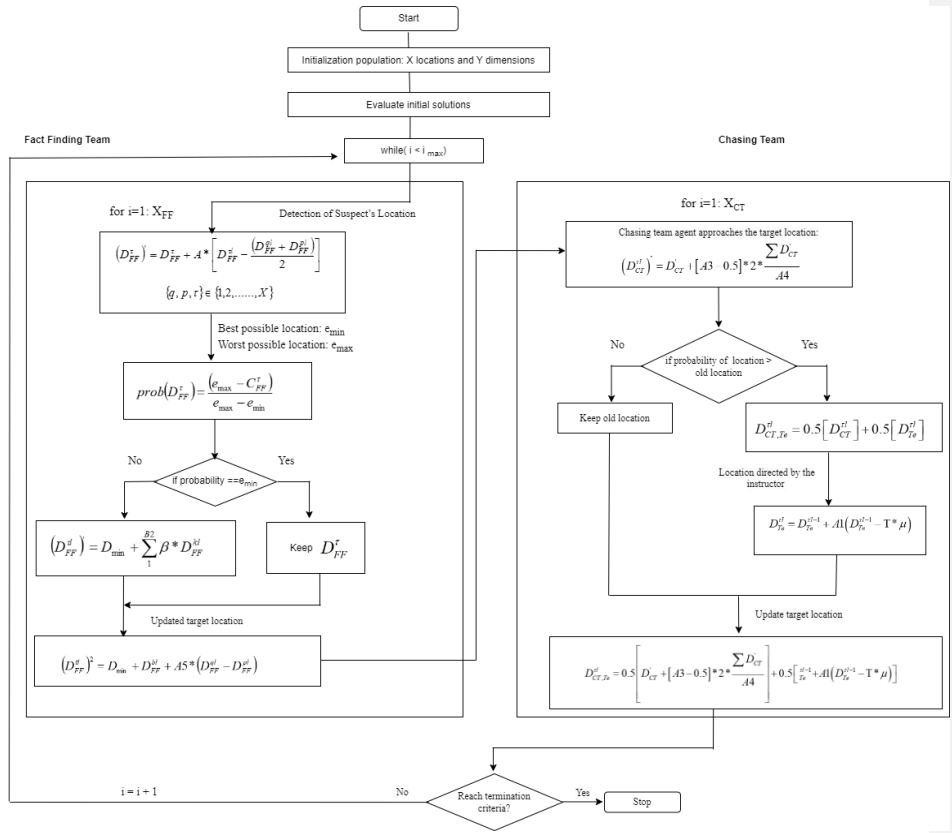

Fig.3: Mathematical model of the FFI algorithm

**K-means clustering:**

K-means clustering is a technique developed based on unsupervised learning, in which the clusters of the given input are employed by considering the centroids (Sinaga et al., 2020). The aim of performing the clustering is to group the input feature vector; the number of the group is represented by the variable K, in which the grouping is employed based on the similarities among the input features (Sinaga et al., 2020, Bagirov et al., 2011).

In the pursuit of enhancing the clustering-based Blood Pressure estimation method, this research has undergone a transition in its approach, opting for the adoption of the Incremental K-means algorithm (Pham et al., 2004, Lin et al., 2004). The utilization of Incremental K-means introduces several compelling advantages that address key challenges encountered in traditional K-means clustering (Bagirov et al., 2011). By harnessing the power of incremental learning and real-time data updates, Incremental K-means brings forth new opportunities to achieve more efficient and accurate BP estimation.

The goal of the incremental k-means algorithm is to minimize the within-cluster sum of squares, which represents the sum of the squared distances between each feature and its assigned cluster center.

**Advantages of incremental K-means algorithm over traditional K-means algorithm:**

1. Real-time Updates: Incremental K-means allows for real-time updates as new data points arrive. It can efficiently handle data streams, making it suitable for dynamic and evolving datasets where data points are continuously added or updated. (Bagirov et al., 2011)

2. Memory Efficiency: Traditional K-means requires storing all data points in memory to calculate the centroids in each iteration. In contrast, incremental K-means updates the centroids on-the-fly, without the need to store the entire dataset in memory. This makes it more memory-efficient, especially for large datasets.

3. Faster Convergence: Incremental K-means can converge faster than traditional K-means, as it updates the centroids incrementally with each new data point. This reduces the number of iterations needed to reach convergence.

4. Scalability: Incremental K-means is more scalable than traditional K-means, particularly for datasets with many data points or dimensions. It can process new data points efficiently without reprocessing the entire dataset (Pham et al., 2004).

5. Online Learning: Incremental K-means supports online learning, where the algorithm continuously learns from new data points, allowing for adaptive and up-to-date clustering models (Hong et al., 2008).

6. Handling Concept Drift: Incremental K-means is well-suited for handling concept drift, which refers to changes in the underlying data distribution over time. It can adapt to changes in the data distribution and update the clustering accordingly (Pham et al., 2004).

7. Flexibility: Incremental K-means allows users to control the rate of incremental updates, making it more flexible in adapting to specific application requirements (Pham et al., 2004).

8. Reduced Computation: As incremental K-means only updates the relevant clusters affected by the new data point, it reduces unnecessary computations compared to traditional K-means, which recalculates all centroids in each iteration.

Overall, incremental K-means is particularly advantageous for scenarios where the data is continuously changing, and real-time updates and scalability are essential. It is well-suited for applications such as online learning, stream data analysis, and handling large and dynamic datasets (Pham et al., 2004, Lin et al., 2004). The pseudocode of incremental clustering algorithm is given in Algorithm 5.1.

**Algorithm 5.1: Incremental K-means clustering algorithm:**
**Input:**
- K: Number of clusters
- data: Input data points
- centroids: Initial cluster centroids
**Output:**
- final_centroids: Updated cluster centroids
- clusters: Assigned clusters for each data point

**Step 1:** Initialize cluster centroids with provided centroids or randomly select K data points as initial centroids.
**Step 2:** Assign each data point to the nearest centroid based on Euclidean distance. This step forms initial clusters.
**Step 3:** Update the centroids by calculating the mean of data points in each cluster.
**Step 4:** Repeat steps 2 and 3 until convergence or a specified number of iterations. Convergence can be determined by checking if the centroids remain unchanged between consecutive iterations.

**Incremental K-means:**
**Step 5:** For each new data point (new_data) in the dataset:
a. Find the nearest centroid (nearest_centroid) to the new_data based on Euclidean distance.
b. Update the centroid (nearest_centroid) by considering the new_data.
To do this, calculate the new mean of data points in the nearest_centroid cluster by incorporating the new_data. The updated centroid can be computed as:

$$\text{nearest\_centroid} = \frac{\text{current\_sum\_of\_data\_points\_in\_nearest\_centroid} + \text{new\_data}}{\text{current\_number\_of\_data\_points\_in\_nearest\_centroid} + 1} \quad (11)$$

c. Reassign the new_data to the nearest_centroid.
d. Repeat steps 2 and 3 with the updated centroid.
**Step 6:** Continue the incremental process as new data points arrive.

**Final Output:**
- final_centroids: Updated cluster centroids after incremental updates.
- clusters: Assigned clusters for each data point after incremental updates.

Let us consider the input feature vector for the k-means clustering $Q$ and the cluster centers are notated as $a1, a2, a3, ..... ak$. Here, from the total features concerning the patients, BP is subcategorized into 3 groups by considering the similarity among the extracted features. Here, by considering each patient's features $Q$, the following functions take place.

Step 1: Initially, the features of patients are assigned to clusters, and the similarity between the assigned features and the cluster centers is evaluated. If the difference is substantial, the patient's features are assigned to a new cluster.

Step 2: The evaluation of the new cluster using its center involves calculating the Euclidean distance, given by:

$$E_{i,j} = \sqrt{\sum_{k=1}^{n} (f_{i,k} - c_{j,k})^2} \qquad (12)$$

Where $E_{i,j}$ is the Euclidean distance between the patient's features $f_{i,k}$ belonging to cluster $i$ and the center $c_{j,k}$ of cluster $j$.

Step 3: These steps are iteratively performed to minimize the loss through Euclidean distance-based evaluation, which can be formulated as:

$$L = \sum_{i=1}^{n} \sum_{j=1}^{k} d_{i,j} \cdot E_{i,j}^2 \qquad (13)$$

Where N is the number of subjects, K is the number of clusters, $d_{i,j}$ is a binary variable indicating if subject i is assigned to cluster j, and $E_{i,j}$ is the Euclidean distance as calculated in Step2.

**Step 4:** The iterations continue until termination criteria are met, typically when the cluster centers cease to change.

For BP estimation, the input feature vector encompasses various attributes extracted from subject's data, like amplitude, frequency, width, and statistical measures. This information-rich feature set characterizes patient data, reflecting relevant factors linked to their BP levels.

With Incremental K-means, the algorithm continuously adapts to evolving feature vectors and updates data point-cluster assignments (Pham et al., 2004, Lin et al., 2004). This ongoing learning approach accommodates real-time updates and changes in data patterns, enabling responsive tracking and categorization of BP levels based on corresponding patient features.

Leveraging Incremental K-means enhances the efficiency and precision of BP estimation. This algorithm systematically organizes subjects with analogous feature patterns into clusters, providing a flexible and dynamic strategy for estimating BP levels (Kumalasari et al. 2020). This data-centric methodology ensures robust and up-to-date subject clustering, contributing to more accurate BP estimation and informed healthcare decisions.

**Combining Clustered Approaches for Blood Pressure Classification:**

During the integration phase, the clusters derived from both the proposed Fact-Finding Instructor Optimization algorithm and the k-means algorithm are merged through a multiplication process. This amalgamation aims to blend the distinct information and attributes obtained from each algorithm, culminating in an improved and accurate Blood Pressure (BP) estimation mechanism.

Upon multiplication of the clusters, BP estimation is executed based on three distinct criteria: low BP, normal BP, and high BP. These criteria delineate the various ranges or classifications of Blood Pressure levels. By associating each cluster with one of these criteria, the estimation process categorizes patients' BP levels into these predefined groups.

The low BP category identifies patients with BP levels below the typical range, signifying hypotension. The normal BP category encompasses patients with BP levels within the healthy range. Meanwhile, the high BP category pertains to patients with BP levels surpassing the standard range, indicating hypertension.

By leveraging these three criteria, the BP estimation process yields a comprehensive evaluation of patients' BP levels, facilitating enhanced identification and classification of their BP statuses. This information holds value for disease diagnosis, ongoing monitoring, and effective management. It empowers healthcare practitioners to make well-informed choices and offer targeted interventions tailored to patients' varying BP conditions.

To conclude, the fusion of Incremental K-means into our clustering-oriented BP estimation framework equips us with the capacity to harness real-time data updates, adapt to dynamic subject features, and ultimately provide refined and dependable BP estimations.

**Dataset description:**

The real-time dataset used in this study consists of Blood Pressure values and corresponding voice signals collected from 25 participants. The participants selected for the dataset fall within the age group of 20 to 65. Out of 25, there are 12 men and 13 female participants, and it is discovered that Blood Pressure has an impact on seven of the participants. By examining patient auditory recordings, the dataset seeks to let doctors precisely determine each patient's Blood Pressure. Each participant's data in the dataset includes their Blood Pressure measurement, which typically consists of Systolic and Diastolic values, and the corresponding voice signal captured during the measurement process. The voice signals are obtained using suitable recording devices or systems.

By utilizing this real-time dataset, researchers and physicians can analyse the relationship between the voice signals and Blood Pressure values. This analysis can aid in developing techniques and algorithms for accurately estimating Blood Pressure based on audio signals.

Table 1: Real-time dataset design types and objectives

| Design Types | Study objective |
|---|---|
| Sample | Homo sapiens |
| Factor | Diagnosis |
| Measurement | Blood Pressure analysis |
| Devices | Boya BY-M1 Omni directional microphone, Omron hem 7120 Blood Pressure monitor |

Table 1 provides an overview of the design types and objectives associated with the real-time dataset used in the study.

**Results and Performance analysis:**

The evaluation of the introduced FFI-driven clustering method's effectiveness in distinguishing between normal and abnormal input audio signals centers on the consideration of epoch size. By scrutinizing the performance metrics across various epoch sizes, valuable insights can be garnered regarding the most suitable epoch duration that maximizes accuracy and overall efficacy in discerning normal and abnormal audio signals through the employed FFI-driven clustering technique. This examination provides a clearer understanding of how the choice of epoch size influences the technique's performance and aids in optimizing its application for audio signal classification.

**Based on the size of the epoch:**

The evaluation is conducted with a training percentage varying from 40 to 90 and on different epoch sizes, ranging from 10 to 50. The assessment of performance encompasses a range of metrics, including the Davies Bouldin score, homogeneity score, completeness score, Jacquard similarity score, silhouette score, and Dunn's index.

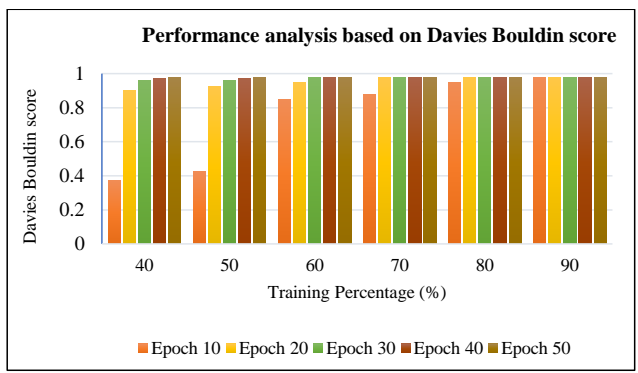

(a)

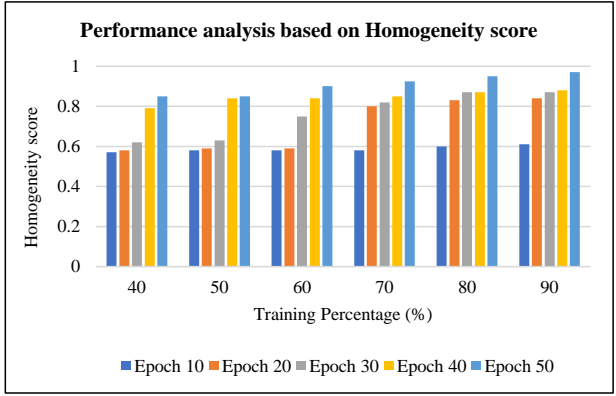

(b)

The Davies Bouldin score, indicative of clustering quality, remains consistently elevated at 0.98. This observation underscores the commendable clustering efficacy of the technique. As the epoch size expands, noticeable enhancements become apparent in key metrics. Notably, the homogeneity, completeness, and Jacquard similarity scores demonstrate a positive trend, reflecting improved precision in clustering accuracy.

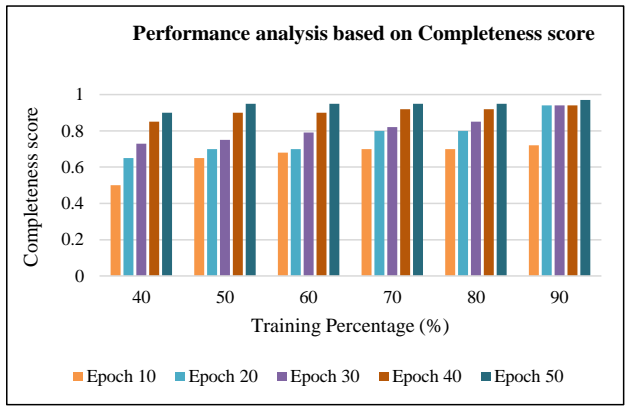

(c)

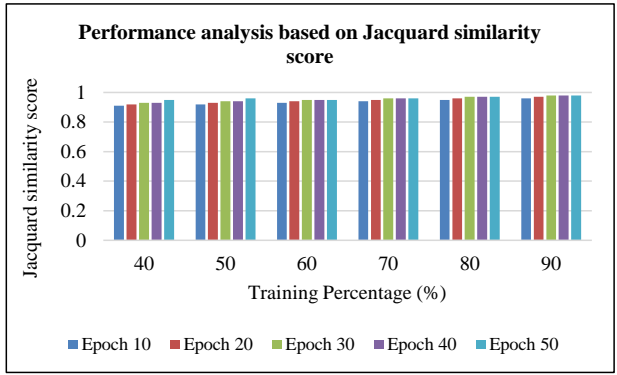

(d)

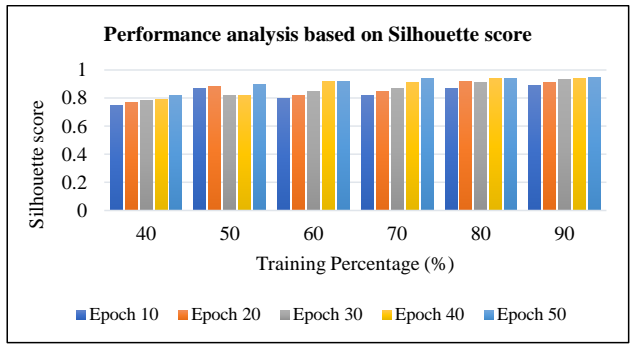

(e)

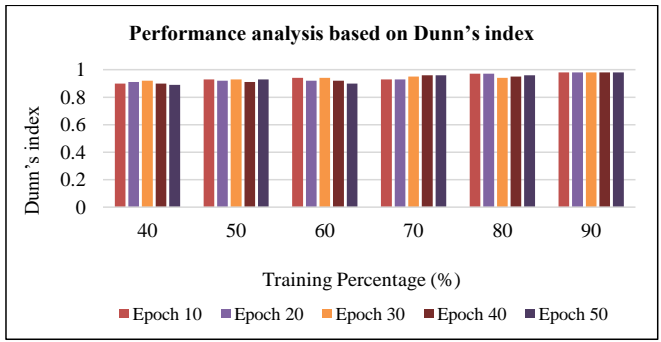

(f)

**Fig. 4:** Performance analysis based on epoch a) Davies Bouldin score.
b) Homogeneity score c) Completeness score d) Jacquard similarity score
e) Silhouette score f) Dunn's index.

Further insights manifest through the silhouette score. These metric gauges the compactness and distinctiveness of clusters. A consistent upward trajectory as the epoch size increases hints at progressively refined clustering differentiation. Equally compelling is the sustained high value of Dunn's index across epochs. This index quantifies the separation of clusters, reaffirming the technique's stable and strong clustering performance. Fig. 4 represents the performance of the FFI-based clustering technique for the various sizes of the epochs.

**Comparative methods:**

In this study, the performance of the proposed FFI-based clustering technique for audio signal analysis is compared with several traditional methods. These methods include:

1. Fuzzy C-means clustering (FCM): Haque and Kim (2013) proposed the FCM clustering algorithm for audio signal analysis. It is based on fuzzy logic and aims to partition the input audio signals into different clusters based on similarity.
2. K-means clustering: Kumalasari et al. (2020) utilized the K-means clustering algorithm for audio signal analysis. K-means is a popular unsupervised clustering algorithm that aims to partition data into K clusters based on minimizing the sum of squared distances between data points and cluster centroids.
3. Agglomerative clustering: Liu et al. (2021) employed the agglomerative clustering algorithm for audio signal analysis. Agglomerative clustering starts with each data point as a separate cluster and progressively merges clusters based on a specific criterion until a desired number of clusters is achieved.
4. Particle Swarm Optimization (PSO)-based clustering: Cai et al. (2020) utilized PSO optimization for clustering audio signals. PSO is an optimization algorithm inspired by the social behaviour of bird flocking or fish schooling, and it aims to find the optimal cluster centroids.
5. Harris Hawk's optimization (HHO)-based clustering: Nagarajan (2020) proposed the HHO optimization algorithm for clustering audio signals. HHO is a nature-inspired optimization algorithm inspired by the hunting behaviour of Harris hawks.
6. Fact Finding Instructor Optimization (FFIO)-based clustering: The FFIO algorithm was developed for clustering audio signals (Chou & Nguyen, 2020). It combines the investigative skills of a fact finder and the knowledge of an instructor to optimize the clustering process.
7. Teaching Learning Optimization (TLO)-based clustering: TLO is an optimization algorithm based on the teaching and learning process, where each solution (student) learns from the best solution (teacher) to improve its performance (Rao et al., 2016).

These traditional methods serve as benchmark approaches for evaluating the performance of the proposed FFI-based clustering technique.

**Comparative analysis:**

Fig. 5 compares the proposed FFI-based clustering technique to various conventional clustering-based strategies in a comparative analysis. The Davies Bouldin score for the FFI-based clustering methodology and the other conventional methods are shown in Fig. 5(a). At a training percentage of 90, the suggested method outperforms the existing TLO-based clustering strategy by 0.20 percent.

The homogeneity score for the FFI-based clustering strategy in comparison to the other conventional methods is shown in Fig. 5(b). Comparing the suggested method to the existing TLO-based clustering strategy at a training percentage of 90, the performance improvement is 6.56%.

The completeness score for the FFI-based clustering methodology and the other conventional methods are shown in Fig. 5(c). Comparing the suggested method to the existing TLO-based clustering strategy at a training percentage of 90, the performance improvement is 0.62%.

The Jacquard similarity score for the FFI-based clustering methodology and the other conventional methods are shown in Fig. 5(d). At a training percentage of 90, the suggested method outperforms the existing TLO-based clustering strategy by 0.10 percent.

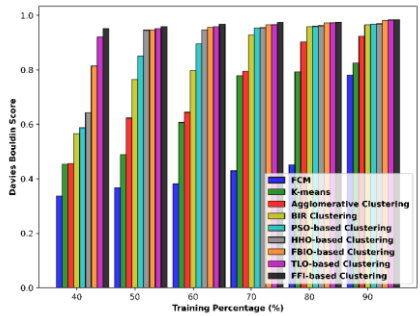

(a)

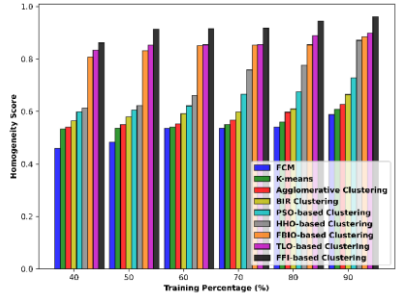

(b)

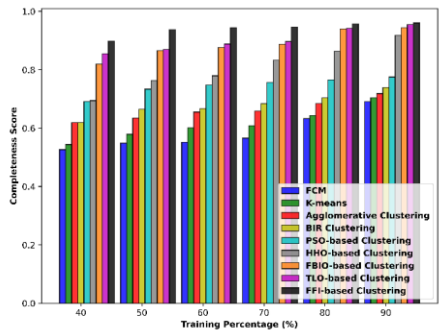

(c)

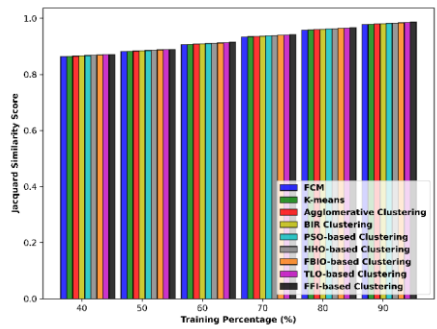

(d)

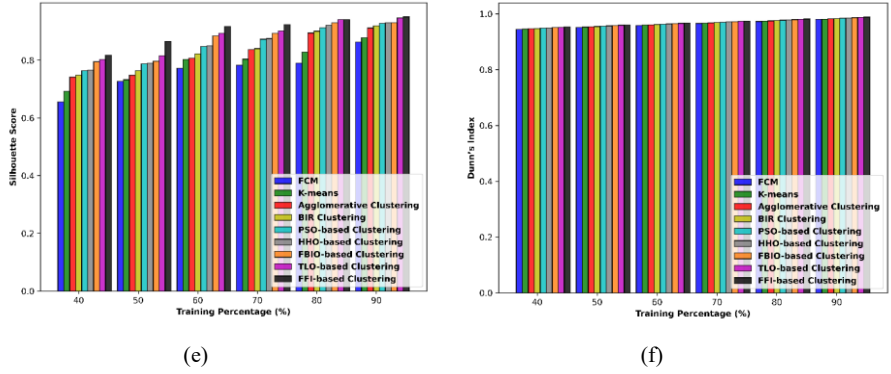

|  (e)  |  (f)  |

**Fig. 5:** Comparative analysis a) Davies Bouldin score b) Homogeneity score c) Completeness score d) Jacquard similarity score e) Silhouette score f) Dunn's index

The Silhouette score for the FFI-based clustering methodology and the other conventional methods are shown in Fig. 5(e). Comparing the suggested method to the existing TLO-based clustering strategy at a training percentage of 90, the performance improvement is 0.53%.

For the FFI-based clustering methodology and the other conventional methods, Fig. 5(f) shows Dunn's index. At a training percentage of 90, the suggested method outperforms the existing TLO-based clustering strategy by 0.10 percent.

**Comparative discussion**

The FFI-based clustering technique consistently shows improved performance compared to the other traditional methods across multiple performance measures. It achieves higher scores in most of the measures, indicating its effectiveness in clustering and analysing audio signals for Blood Pressure estimation.

1. **FFI Outperforms Traditional Methods:** The FFI-based clustering technique achieves the highest scores in terms of Davies Bouldin score, homogeneity score, completeness score, Jacquard similarity score, silhouette score, and Dunn's index. This demonstrates its superiority in capturing the underlying patterns and grouping the input audio signals accurately.
2. **Strong Homogeneity and Completeness:** The FFI-based clustering technique stands out in terms of homogeneity and completeness scores. It achieves a homogeneity score of 0.961 and completeness score of 0.961, indicating high consistency within the clusters and capturing all the data points within the respective clusters effectively.
3. **Consistent Improvement:** Compared to the previous TLO-based clustering technique, the FFI-based clustering technique shows notable performance improvements across different measures. The performance improvement ranges from 0.10% to 6.56%, further highlighting the efficacy of the FFI-based approach.

**Time Series data analysis:**

Time series data refers to a collection of observations recorded over regular time intervals, forming a sequence of data points(Pham et al., 2004, Lin et al., 2004) . In our study, time series voice data captures the speech signals of different individuals at various time points, allowing us to examine how speech characteristics affect the individual Blood Pressure.

**Case Study-You Tube Videos:**

In the digital age, YouTube has emerged as one of the most popular platforms for content consumption, providing a vast array of videos ranging from entertainment and educational content to news and lifestyle vlogs. With billions of users and an ever-expanding library of videos, YouTube has become a powerful medium of communication, shaping the way we access information and experience emotions. Amidst this vast sea of diverse content, it is fascinating to explore the potential impact that YouTube videos can have on our emotional well-being and physiological responses.

In recent years, there has been a growing interest in understanding how various forms of media, including videos, can influence human emotions and even physiological parameters such as Blood Pressure (BP). Studies have shown that emotional experiences can significantly affect Blood Pressure levels, with heightened emotions often leading to temporary fluctuations in BP.

In this context, exploring the relationship between YouTube videos and their impact on Blood Pressure assumes great significance. YouTube's vast repository of content covers a spectrum of emotions, from heartwarming and humorous videos that elicit joy and laughter to intense and suspenseful videos that evoke fear or anxiety. The emotional content and intensity conveyed through these videos have the potential to elicit diverse physiological responses in viewers.

The research work aims to present an efficient BP estimation technique using time series speech data extracted from YouTube videos. The approach capitalizes on the Fact-Finding Instructor Optimization algorithm, designed to handle time series data, and leverages both YouTube data's abundance and real-time dataset's ground truth labelling.

The proposed time series-based BP estimation method begins with preprocessing the YouTube speech data to remove noise and artifacts using adaptive filters. Feature extraction is performed to generate a feature vector, comprising statistical-based features (e.g., zero crossing, entropy) and formant features, tailored for time series analysis to reduce computational complexity. Subsequently, k-means clustering, and the Fact-Finding Instructor Optimization algorithm are applied to group similar time series feature vectors together. To train the model, the real-time dataset with actual Blood Pressure recordings is used, providing ground truth labels for model optimization. The model is trained on the real-time dataset to enhance BP estimation accuracy. The clustering algorithm is fine-tuned with both YouTube time series speech data and the real-time dataset, taking advantage of both dataset's characteristics.

**Time Series Analysis of Emotional States in Daily Videos:**

To enhance the BP estimation accuracy, the model is trained using a real-time dataset with actual BP recordings as ground truth labels. The successful integration of both online video data and the real-time dataset demonstrates the potential of this time series-based clustering approach for improving BP estimation in medical research and healthcare applications.

The proposed method integrates a newly developed clustering algorithm, combining k-means clustering with the Fact-Finding Instructor Optimization algorithm to effectively capture temporal patterns in the speech signals, leading to enhanced BP estimation accuracy. By considering YouTube speech data from the daily online videos in a time series format, this research aims to leverage the rich temporal information embedded in the data, furthering the understanding of BP variations.

**Methodology:**

**Data Collection:** A collection of daily videos featuring some cool, happy, motivational thoughts like Spiritual Gurus as well as some aggressiveness, was obtained from online sources.

**Daily Emotional Segments:** The time series analysis of the daily online videos commences with a preprocessing step, wherein adaptive filters are utilized to remove noise and artifacts, ensuring the quality of the speech signals for subsequent analysis. Feature extraction is then carried out to generate a feature vector comprising both statistical-based features (e.g., zero crossing, entropy) and formant features, specifically tailored to facilitate time series analysis and reduce computational complexity (Lin et al., 2004)

Here are the results of extracted features from time series analysis of angry and calm audio clips:

1. **Feature Extraction:** Several relevant features from both angry and calm audio clips are extracted. These features included the Zero Crossing Rate, Spectral Centroid, Energy, and the first three Mel-Frequency Cepstral Coefficients (MFCC1, MFCC2, and MFCC3). Additionally, we obtained the first two Formant Frequencies (Formant 1 and Formant 2) to capture distinct characteristics of the vocal tract.
2. **Distinct Patterns:** The analysis revealed distinct patterns in the extracted features between angry and calm audio clips. Notably, the Zero Crossing Rate, Spectral Centroid, and Energy tended to show significant variations between the two emotional states. Calm audio clips generally exhibited lower Zero Crossing Rate, lower Spectral Centroid, and lower Energy values compared to the corresponding features in angry audio clips.

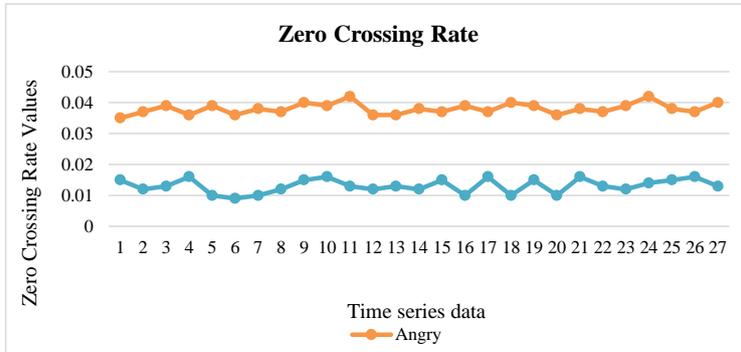

Fig. 6: Zero Crossing Rate values of angry and calm audio signals

3. **MFCC Differences:** The Mel-Frequency Cepstral Coefficients (MFCCs) also exhibited variations between the two emotional states. In particular, the MFCC1 coefficient, which represents the overall energy level, tended to be lower in calm audio clips compared to angry ones. Additionally, certain other MFCCs (e.g., MFCC2 and MFCC3) displayed differences in their spectral characteristics.

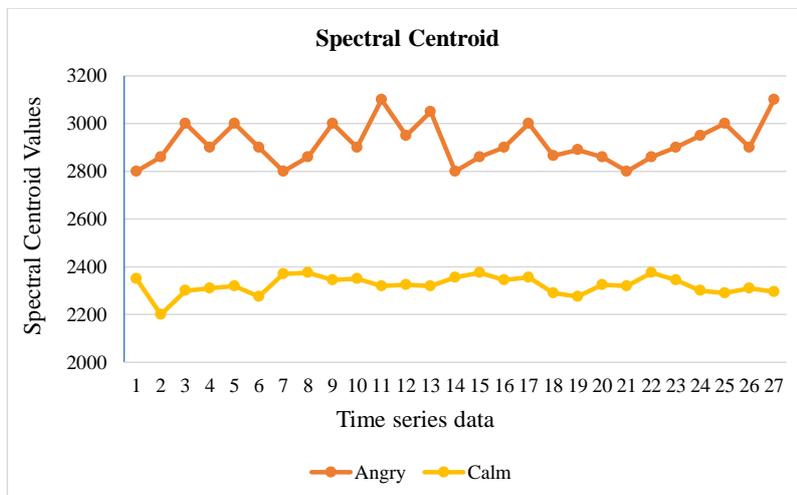

Fig.7: Spectral Centroid values of angry and calm audio signals

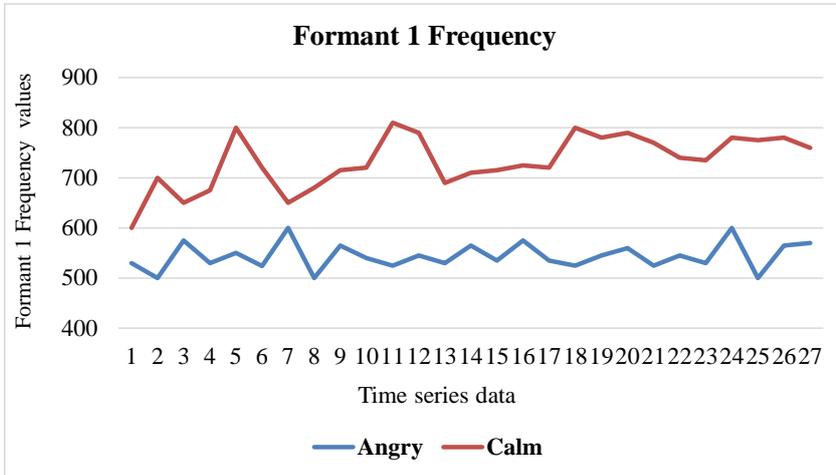

Fig.8: Formant 1 Frequency values of angry and calm audio signals

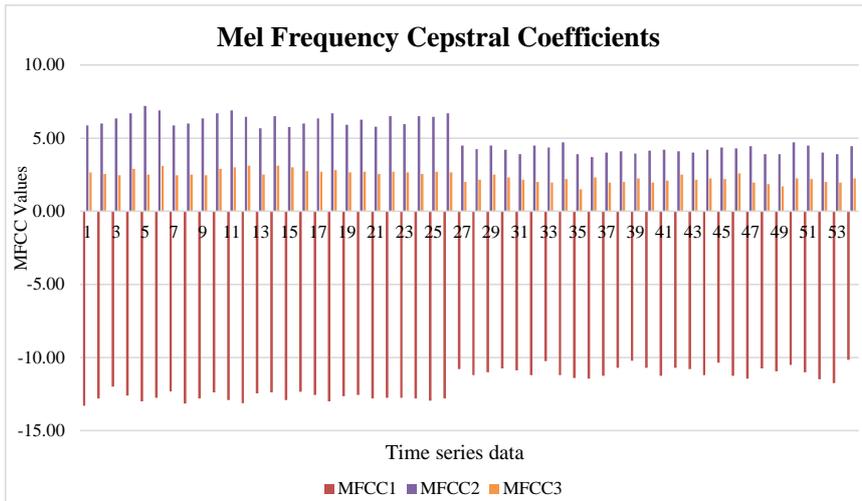

Fig.9: MFCC values of angry and calm audio signals

4. **Formant Frequencies:** The Formant 1 and Formant 2 frequencies showed distinctive patterns between angry and calm audio clips. These frequencies are related to the resonance of the vocal tract and were found to vary significantly depending on the emotional expression. The results are visualized in Fig.6 to Fig.9.

To cluster the time series, feature vectors effectively, we employ a two-pronged approach. Initially, k-means clustering is applied to group similar feature vectors. Subsequently, the Fact-Finding Instructor Optimization algorithm is integrated into the clustering process, further enhancing the accuracy of the BP estimation. To optimize the model's performance, the clustering algorithm is fine-tuned using a real-time dataset with actual BP recordings, where ground truth labels are available for training.

**Results and Discussion:** The successful integration of both the daily online videos in time series format and the real-time dataset leads to improved BP estimation accuracy. The novel clustering algorithm, incorporating both k-means clustering and the Fact-Finding Instructor Optimization algorithm, effectively captures temporal patterns in the speech signals, contributing to the overall effectiveness of the proposed technique. The regression statistics provide valuable insights into the relationship between the Blood Pressure and the feature vector. Let's interpret the key regression statistics depicted in Table 2:

| Table 2: Regression Statistics | |
|---|---|
| Multiple R | 0.89418 |
| R Square | 0.9234 |
| Adjusted R Square | 0.743879 |
| Standard Error | 4.339477 |

1. **Multiple R:** The multiple correlation coefficient (Multiple R) is a measure of the overall correlation between the independent variables and the dependent variable. In this case, the Multiple R value is approximately 0.894, indicating a strong positive relationship between the feature vector and the Blood Pressure.
2. **R Square:** The coefficient of determination (R Square) represents the proportion of the variance in the dependent variable that can be explained by the independent variables. Here, the R Square is approximately 0.92, which means that approximately 92% of the variability in the Blood Pressure is accounted for by the feature vector in the model.
3. **Adjusted R Square:** The adjusted R Square considers the number of independent variables and adjusts the R Square accordingly. In this case, the Adjusted R Square is approximately 0.744, indicating that the model is well-fitted to the data, and the chosen feature vector is strongly related to the Blood Pressure.
4. **Standard Error:** The standard error estimates the average difference between the actual values of the dependent variable and the predicted values from the regression model. A lower standard error indicates a better fit of the model to the data. Here, the standard error is approximately 4.339, which means that the model's predictions might have an average error of around 4.339 units from the actual values of the dependent variable.

**Interpretation of Regression Analysis:**

The strong positive multiple R indicates a significant correlation between the independent variables and the dependent variable. The high R Square and Adjusted R Square values suggest that the independent variables in the model are effective predictors, explaining a substantial portion (approximately 92%) of the variability in the dependent variable.

Additionally, the relatively low standard error of approximately 4.339 indicates that the model's predictions are generally close to the actual values of the dependent variable.

Overall, the regression analysis demonstrates a robust relationship between the extracted features from audio and the Blood Pressure. The model's strong performance suggests that the selected features are valuable for predicting the Blood Pressure, leading to accurate estimates or predictions.

**Conclusion**

This research presents a comprehensive time series-based clustering approach for BP estimation using daily online videos. The incorporation of the novel clustering algorithm, fine-tuned with real-time dataset, showcases the adaptability and robustness of the proposed technique. By leveraging temporal patterns in the speech signals, the method offers promising advancements in healthcare maintenance and disease diagnosis in cardiovascular-related applications. Overall, this research highlights the potential of time series analysis in medical data analysis, paving the way for future developments in cardiovascular function evaluation using large-scale speech datasets.

Overall, the FFI-based clustering technique demonstrates superior performance and outperforms the other traditional methods in accurately clustering and analysing audio signals for Blood Pressure estimation. It showcases its potential as an effective method for assisting physicians in determining the Blood Pressure of patients using audio signals.

By integrating the expertise of the instructor and the investigative behaviour of the fact-finding team, the proposed Fact-Finding Instructor Optimization algorithm achieves more accurate solutions for clustering the BP features. The algorithm balances intensification and diversification phases to avoid getting trapped in local optima and ensures the discovery of the global best solution. This leads to optimal clustering of the patient's BP, ultimately contributing to accurate diagnosis.